\documentclass[nopreprintline, 3p, number, sort&compress]{elsarticle}

\usepackage{url}

\usepackage{graphicx}

\usepackage{color}
\usepackage{amsmath}
\usepackage{enumerate}
\usepackage{subfigure}
\usepackage{flushend}
\usepackage{booktabs}
\usepackage{bm}
\usepackage{algorithm}
\usepackage{algpseudocode}
\usepackage{amssymb}
\usepackage{mathtools}
\usepackage{breqn}
\usepackage{threeparttable}
\usepackage{multirow}
\usepackage[OT1]{fontenc}

\begin{document}
	\begin{frontmatter}
		\title{Saturated Non-Monotonic Activation Functions}
		\author{Junjia Chen}
		\author{Zhibin Pan\corref{cor1}}
		\ead{zbpan@xjtu.edu.cn}
		\cortext[cor1]{Corresponding author}
		\address{School of Electronic and Information Engineering, Xi’an Jiaotong University, Xi’an 710049, P. R. China}
		
		\begin{abstract}
			Activation functions are essential to deep learning networks. Popular and versatile activation functions are mostly monotonic functions, some non-monotonic activation functions are being explored and show promising performance. But by introducing non-monotonicity, they also alter the positive input, which is proved to be unnecessary by the success of ReLU and its variants. In this paper, we double down on the non-monotonic activation functions' development and propose the Saturated Gaussian Error Linear Units by combining the characteristics of ReLU and non-monotonic activation functions. We present three new activation functions built with our proposed method: SGELU, SSiLU, and SMish, which are composed of the negative portion of GELU, SiLU, and Mish, respectively, and ReLU's positive portion. The results of image classification experiments on CIFAR-100 indicate that our proposed activation functions are highly effective and outperform state-of-the-art baselines across multiple deep learning architectures.
		\end{abstract}

		\begin{keyword}
			Adaptive activation function, GELU, deep learning
		\end{keyword}
	\end{frontmatter}
	
	\section{Introduction}

	Deep learning has become a widely popular technique that has proven to be highly effective in a variety of applications. The academic and engineering communities have been focusing on this area for years. Thanks to the utilization of general-purpose computing on graphics processing units (GPGPU) \cite{gpgpu_1, gpgpu_2, gpgpu_3}, deep learning techniques have been evolving rapidly with hundreds of new models proposed each year. These methods excel in areas such as computer vision (CV) \cite{nn_gpgpu_3, misc_transformer} and natural language processing (NLP) \cite{misc_attention, dl_nlp_overview}.

	Deep learning networks are composed of thousands of affine and nonlinear transformations, with the nonlinear transformations, called activation functions, mimicking neuron firing mechanisms. Early neural network research used the Heaviside step function as the activation function, but its gradient of 0 makes it unsuitable for gradient-based optimization. Several S-shape functions, like the Sigmoid function and the hyperbolic tangent function (tanh), were proposed to be smooth variants of the Heaviside step function. However, neural networks with such activation functions suffer from the vanishing gradient problem, where the gradients obtained from backpropagation are too small to initiate weight updates.
	
	The Rectified Linear Unit (ReLU)\cite{relu_1,relu_2} is one of the most popular activation functions due to its simplicity, fast convergence speed, and sparsity. However, it suffers from a problem called dying ReLU that causes the weights to stop updating. To solve this issue, variants of ReLU have been proposed, such as LeakyReLU\cite{leakyrelu}, which scales down negative values instead of shutting them down, and Parametric ReLU (PReLU)\cite{prelu}, which parameterizes the LeakyReLU's slope of the negative part, improving the model fitting with nearly zero extra computational cost.
	
\begin{figure}[H]
		\centerline{\includegraphics[width=0.3\columnwidth]{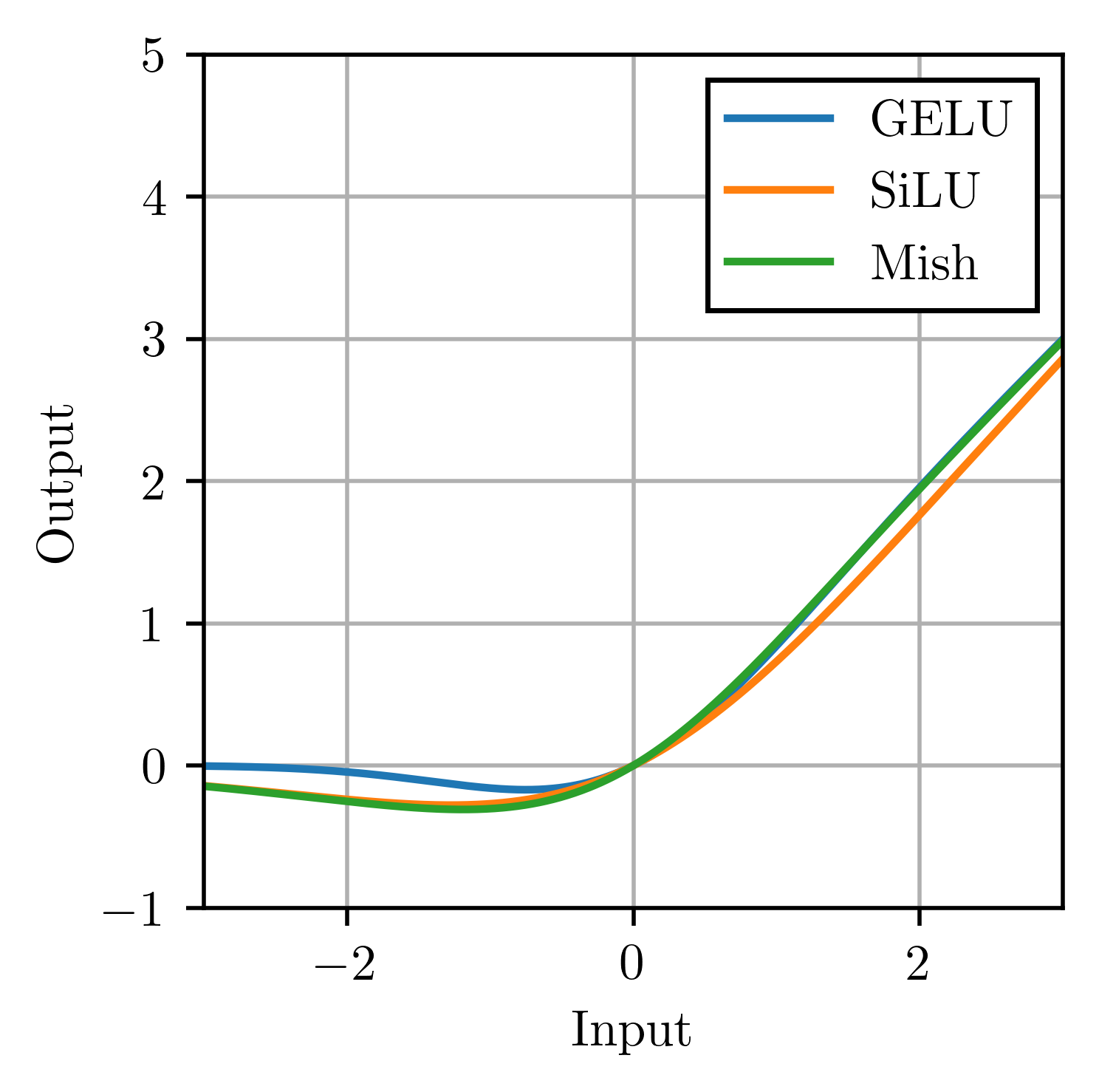}}
		\label{gelu_silu_mish_plots}
		\caption{Non-monotonic activation functions.}
	\end{figure}

	Although most activation functions are monotonic, some non-monotonic activation functions have shown excellent performance. Fig. \ref{gelu_silu_mish_plots} shows plots of some popular non-monotonic activation functions. GELU and SiLU were developed by combining properties from dropout, zoneout, and ReLU \cite{gelu}. Swish\cite{swish} has the same form as SiLU and has a trainable parameter, but it was independently discovered through a meta-learning technique. Mish\cite{mish} was later developed, influenced by Swish, with a smoother loss landscape and better performance. Power Function Linear Unit (PFLU)\cite{pflu} is a non-monotonic activation function that maintains the sparsity of the negative part while introducing negative activation values and non-zero derivative values for the negative part.

	Many popular activation functions are composed of a mixture of multiple basic functions. For example, LeakyReLU consists of two linear functions for the positive and negative parts, respectively. Exponential Linear Unit (ELU) \cite{elu} uses an exponential function for the negative part and a linear function for the positive part. The authors of Swish constructed a search space containing various basic functions such as linear functions, trigonometric functions, and exponential functions. They combined these functions through multiple binary operations to form the optimal activation function Swish.

	Inspired by these works, we observed that many activation functions treat positive and negative values differently. For example, ReLU uses the identity function for the positive part, allowing it to activate positive features without distortion, but it completely discards negative values, leading to the dying ReLU problem. GELU is a non-monotonic function, allowing some negative values to be activated through a non-monotonic region, but it also has a small amount of nonlinearity in the positive region, which causes some distortion in activated positive inputs. Non-monotonic activation functions like GELU have the following form:

	\begin{equation} \label{non_monotonic_ac}
		f(x) = x \cdot s(\beta x),
	\end{equation}
	where $s(\cdot)$ is an activation function with a value range of $[0,1]$, $\beta$ is a hyperparameter or trainable parameter, typically set to $1$ if it is a hyperparameter. The non-monotonic activation functions that we used in this paper include GELU, SiLU, and Mish, they are defined as follows:

	\begin{equation} 
		\begin{aligned} 
			\text{GELU}(x) &= x \frac{1+\text{erf}(\frac{x}{\sqrt{2}})}{2}, \\
			\text{SiLU}(x) &= x \frac{1}{1+e^{-x}}, \\
			\text{Mish}(x) &= x \text{tanh}(\ln(1+e^x)). \\
		\end{aligned} 
	\end{equation}

	These activation functions are non-monotonic in the negative region and can effectively utilize negative input information while introducing nonlinearity to improve network generalization. They also have gate-like properties, where $s(\beta x)$ can be seen as a gate function that controls the forward propagation of $x$, but the value of the gate function depends on $x$. The positive parts of these activation functions are approximately linear, but due to the logarithmic and exponential operations in the expression of these functions, there is some distortion in activated positive inputs. Therefore, we propose to combine ReLU, which activates positive features without distortion, and other activation functions with strong expression capability in the negative region, to obtain performance gains.

	In this paper, we proposed a non-monotonic activation function Saturated Gaussian Error Linear Units (SGELU). 1) It takes advantage of the non-monotonic in the negative part. 2) it keeps the positive part simple and efficient as ReLU and thus has a constant gradient which enables more efficient gradient descent. We evaluate our method and find that our method constantly performs better compared to the most popular activation functions.
	
	\section{Proposed Method}

	\subsection{Constructing Non-Monotonic Activation Functions}
	We combine the positive part of ReLU and the negative part of non-monotonic activation functions to construct new activation functions as follows:
	\begin{equation}
		f_s(x) = 
		\begin{cases}
			x, & x \ge 0 \\
			x \cdot s(\beta x), &x < 0\\
		\end{cases},
	\end{equation}
	Our method can keep the non-monotonic negative part while activating positive neurons losslessly. Based on this method, we propose three activation functions as follows:
	\begin{equation}
		\begin{aligned}
			\text{SGELU}(x) &= \text{max}(x \frac{1+\text{erf}(\frac{x}{\sqrt{2}})}{2}, x) \\
			\text{SSiLU}(x) &= \text{max}(x \frac{1}{1+e^{-x}}, x) \\
			\text{SMish}(x) &= \text{max}(x \text{tanh}(\ln(1+e^x)), x) \\
		\end{aligned}
	\end{equation}

	\begin{figure*}[ht]
		\centering
		\subfigure[]{\includegraphics[width=0.3\columnwidth]{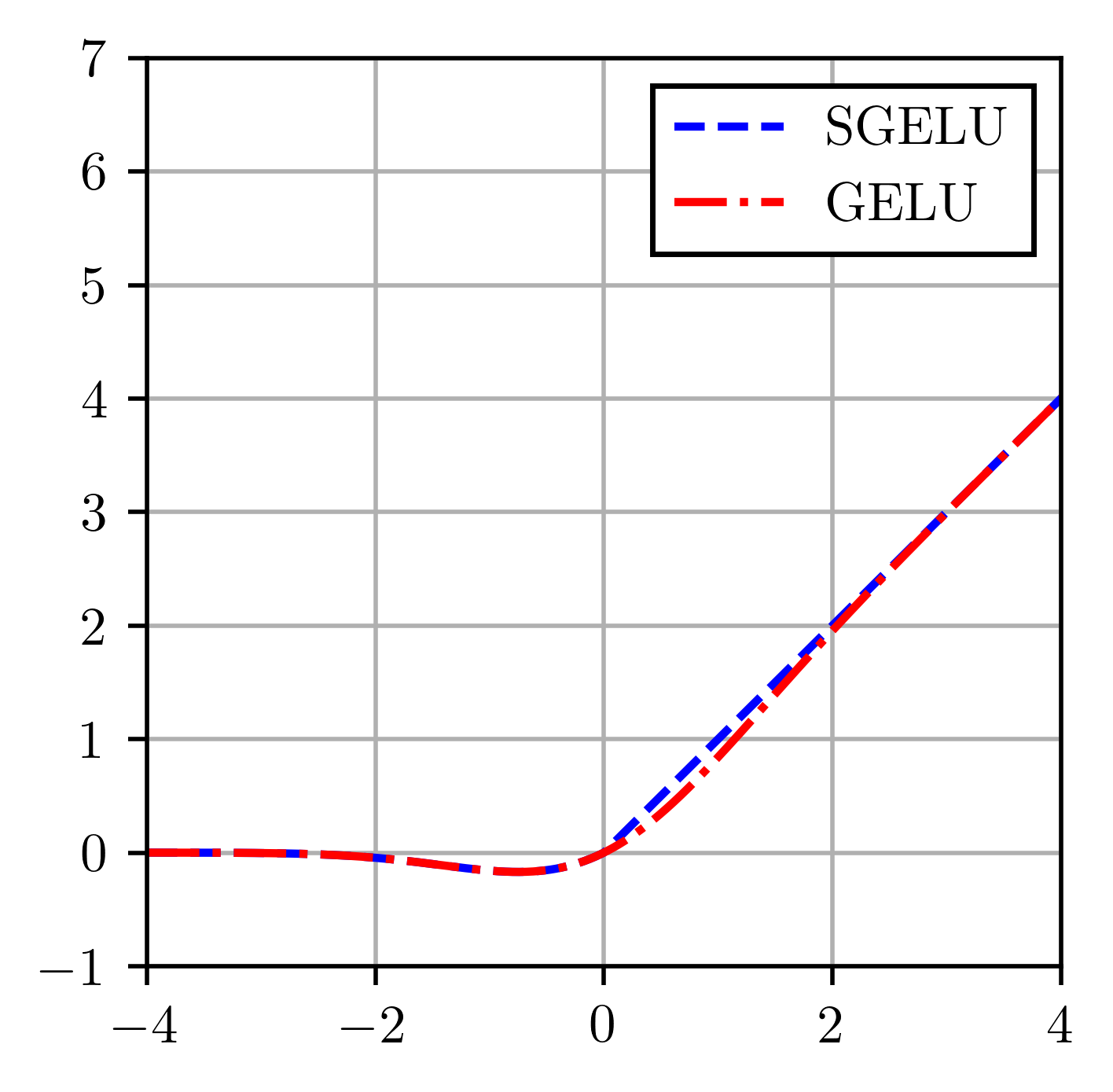}}
		\subfigure[]{\includegraphics[width=0.3\columnwidth]{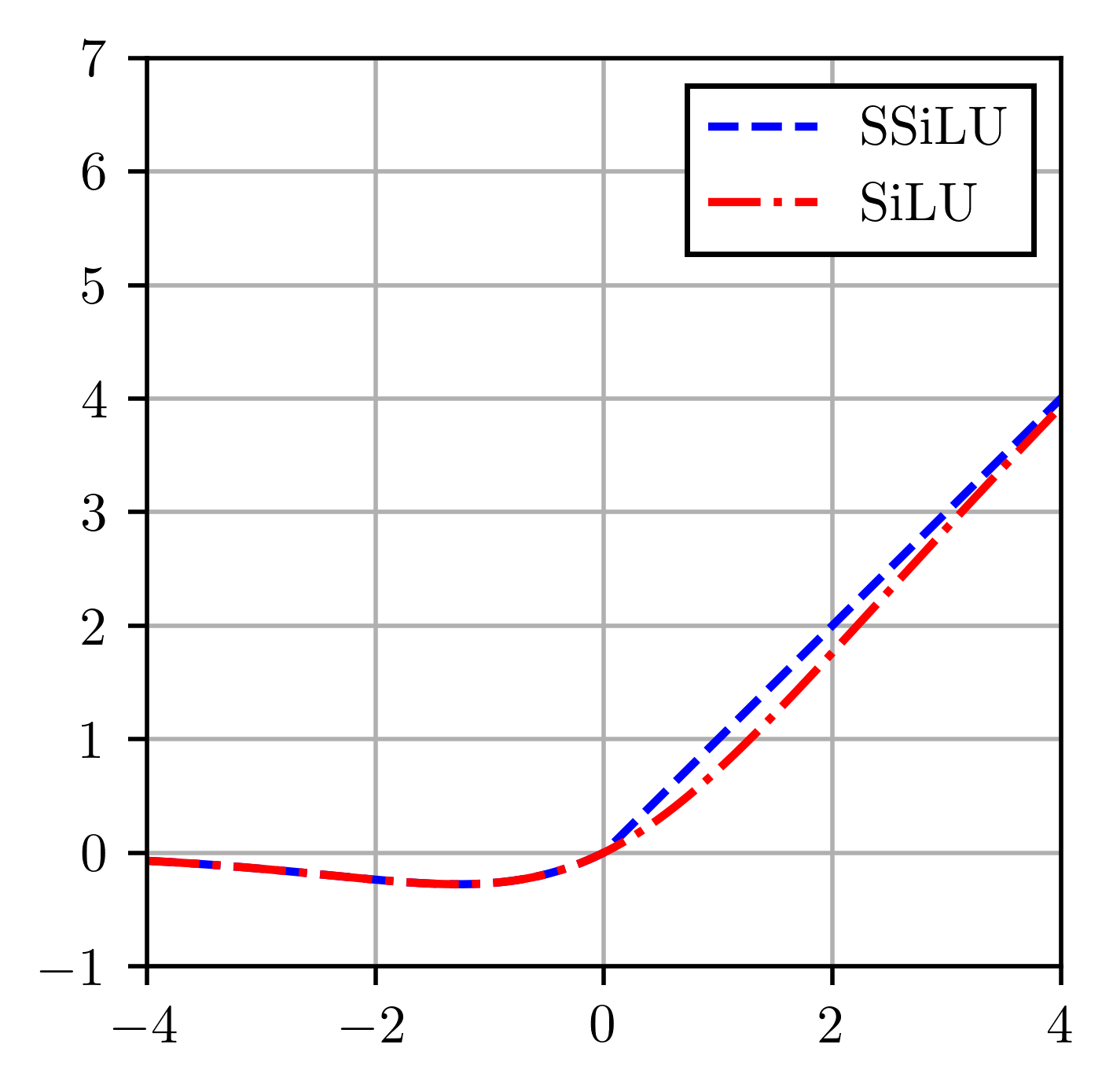}}
		\subfigure[]{\includegraphics[width=0.3\columnwidth]{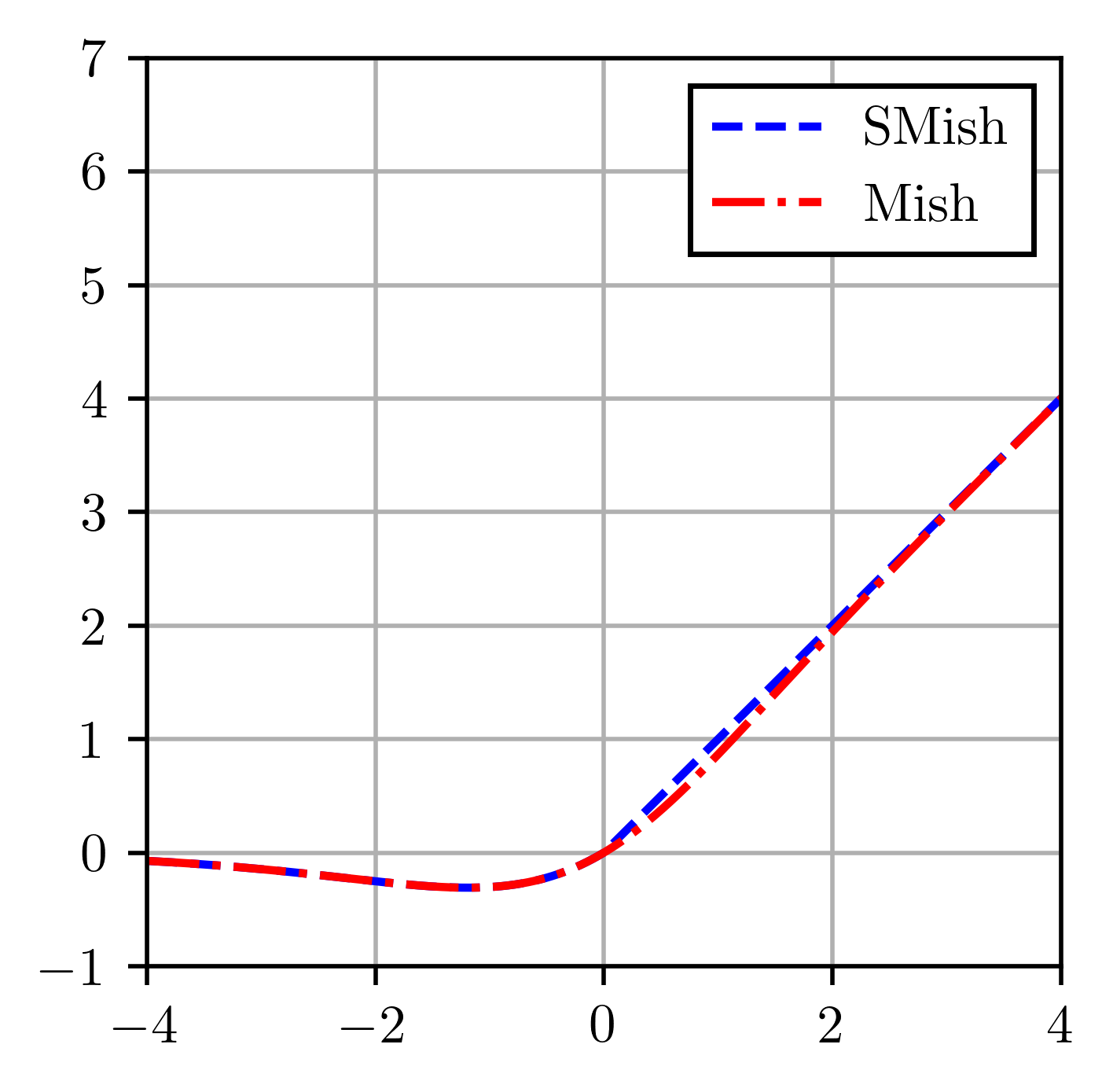}}
		\caption{Plots of our methods and their original non-monotonic activation functions. (a) GELU and SGELU. (b) SiLU and SSiLU. (c) Mish and SMish.}
		\label{sgelu_ssilu_smish}
	\end{figure*}

	\begin{figure*}[ht]
		\centering
		\subfigure[]{\includegraphics[width=0.3\columnwidth]{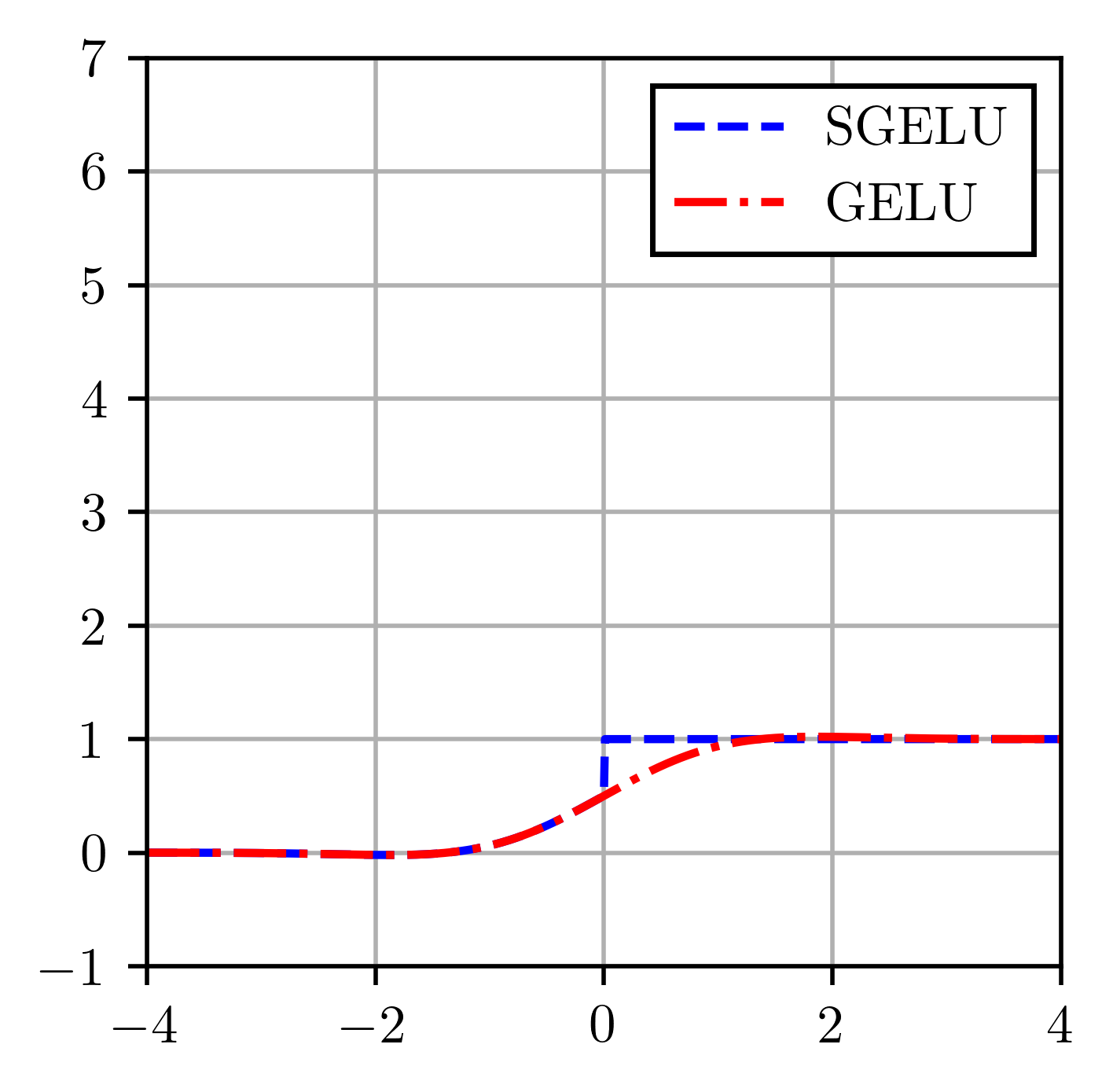}}
		\subfigure[]{\includegraphics[width=0.3\columnwidth]{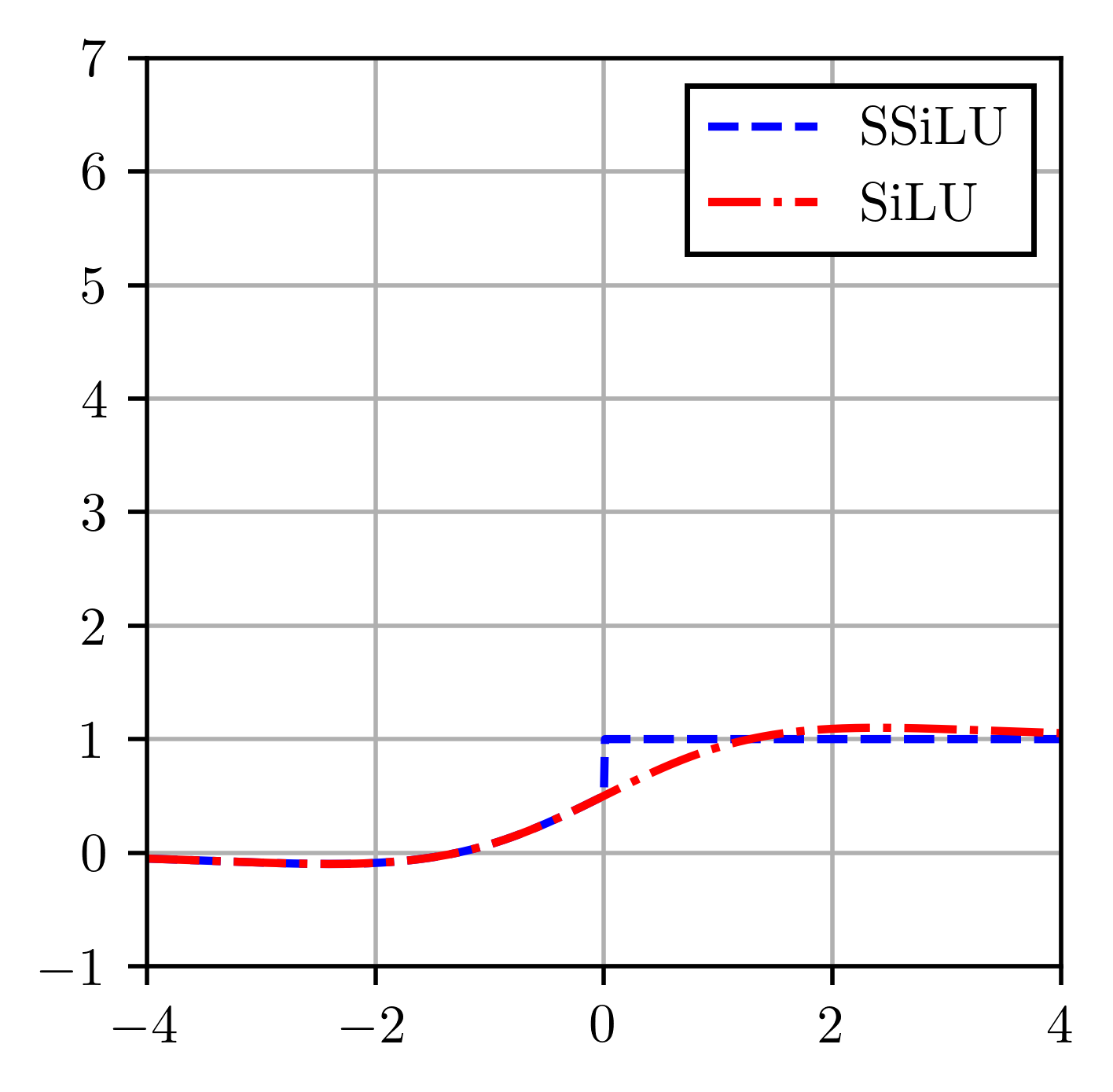}}
		\subfigure[]{\includegraphics[width=0.3\columnwidth]{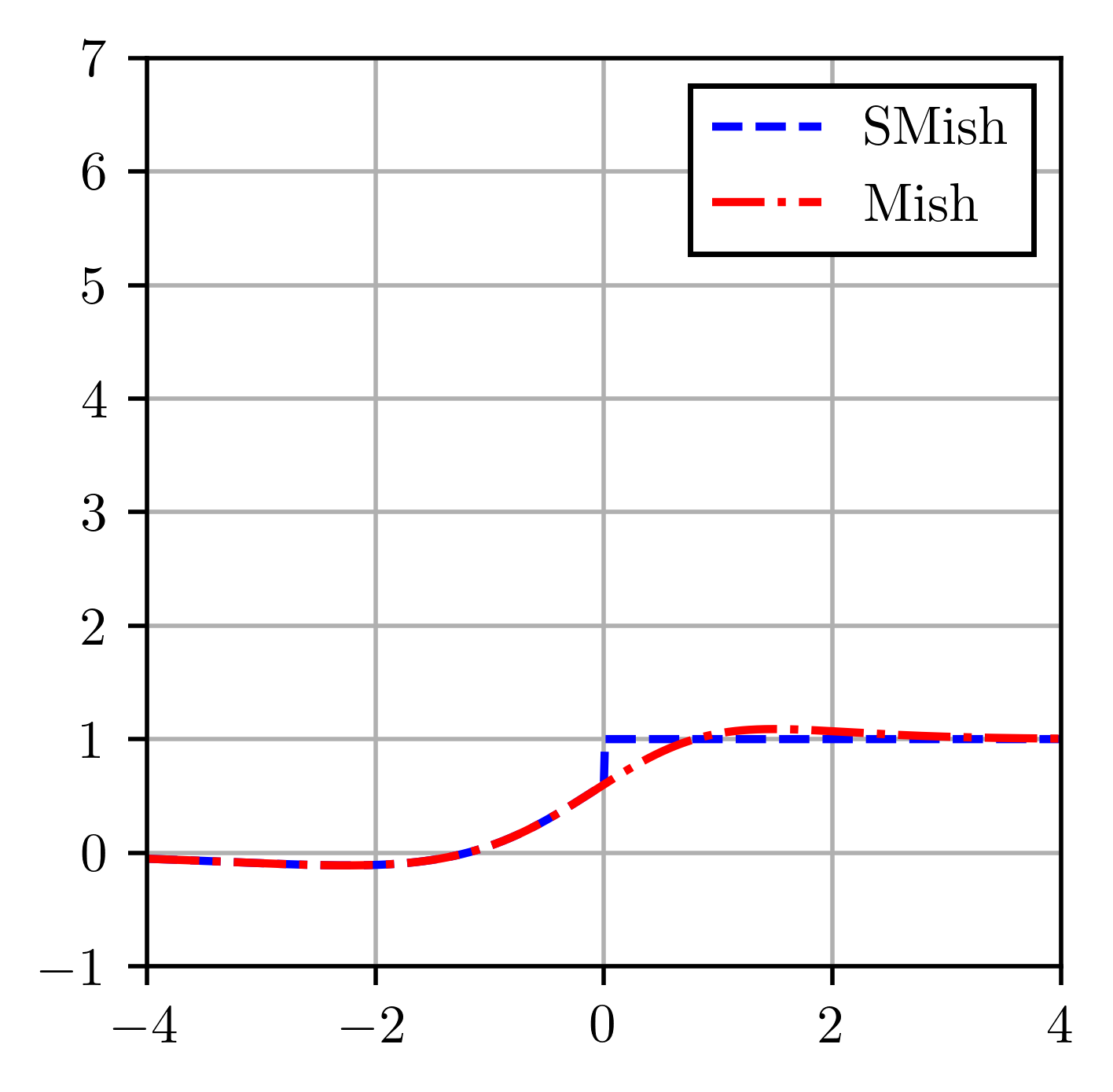}}
		\caption{Plots of the gradients of our methods and their original non-monotonic activation functions. (a) GELU and SGELU. (b) SiLU and SSiLU. (c) Mish and SMish.}
		\label{gradient_sgelu_ssilu_smish}
	\end{figure*}

	The gradients of these activation functions can be derived as follows:
	\begin{equation}
		\begin{aligned}
			\frac{\mathrm{d}\text{SGELU}(x)}{\mathrm{d}x} &= 
			\begin{dcases}
				\frac{x}{2\pi} {\rm e}^{-\frac{x^2}{2}} + \frac{1}{2}[1+{\rm erf}(\frac{x}{\sqrt{2}})], &x<0 \\
				1, &x \ge 0
			\end{dcases} \\
			\frac{\mathrm{d}\text{SSiLU}(x)}{\mathrm{d}x} &= 
			\begin{dcases}
				\frac{1+e^{-x}+xe^{-x}}{(1+e^{-x})^2}, &x<0 \\
				1, &x \ge 0
			\end{dcases} \\
			\frac{\mathrm{d}\text{SMish}(x)}{\mathrm{d}x} &= 
			\begin{dcases}
				\frac{e^x [4(x+1)+4e^{2x}+e^{3x}+e^{x}(4x+6)]}{(2e^x+e^{2x}+2)^2}, &x<0 \\
				1, &x \ge 0
			\end{dcases} \\
		\end{aligned}
	\end{equation}

	The difference between the original non-monotonic nonlinearity and our method is that our method has a higher and constant gradient on the positive part. This enables gradient descent to be more efficient.

	\subsection{Pass Rate Saturation}
	Every non-monotonic activation function described by Eq. \eqref{non_monotonic_ac} can be explained as an expected transformation of a stochastic process that multiplies a random 0-1 mask $m$ with input. The value of random 0-1 mask $m$ depends on the inputs as follows:
	\begin{equation} \label{gelu_m}
		m \sim \text{Bernoulli}(F(x)),
	\end{equation}
	where $F(x)$ is used to control the pass rate depending on the input. The result activation function is the expected transformation of this process:
	\begin{equation}
		\begin{aligned}
		  f(x) &= E(mx) \\
		  &= xF(x) \\
		\end{aligned}
	\end{equation}
	The pass rate functions for GELU, SiLU and Mish are defined as follows:
	\begin{equation}
		\begin{aligned}
			F_{\text{GELU}}(x) &= \frac{1}{2}[1+\text{erf}(\frac{x}{\sqrt{2}})], \\
			F_{\text{SiLU}}(x) &= \frac{1}{1+e^{-x}}, \\
			F_{\text{Mish}}(x) &= \text{tanh}(\ln(1+e^x)). \\
		\end{aligned}
	\end{equation}
	The stochastic process can not guarantee all positive values to be activated, which distorts positive signals.
	ReLU can also be written as $xF(x)$ as follows:
	\begin{equation}
		\begin{aligned}
			\text{ReLU}(x) &= \text{max}(x, 0) \\
			&= x 
			\begin{cases}
				1, &x \ge 0 \\
				0, &x < 0 \\
			\end{cases} \\
			&= x F_{\text{ReLU}}(x).
		\end{aligned}
	\end{equation}
	Where $F_{\text{ReLU}}(x)$ is the pass rate function for ReLU. ReLU has a pass rate of 100\% when inputs are positive, and thus losslessly activates all positive values.

	Our method can be seen as saturating the pass rate of the original non-monotonic activation functions when inputs are positive, which makes the positive values to be activated without distortion.

	\section{Experiments}
	We compare our methods to the most popular activation functions on CIFAR-100 imagine classification task\cite{cifar}. CIFAR-100 is a dataset that has 100 classes, containing 500 training images and 100 test images for each class.  We use MobileNet, MobileNetV2, VGG-11 and VGG-13 networks to evaluate our activation functions. A stochastic gradient descent (SGD) optimizer with a momentum of $0.9$ and a weight decay\cite{weightdecay} of $5 \times 10^{-4}$ is used to train all networks. The learning rate starts at $0.1$ and is divided by $5$ in $50$th, $120$th, and $160$th epochs. The original activation functions for these models are ReLU. To test our method and other baseline activation functions, We simply replaced every activation function in the model with target activation functions.


	\begin{table}[H]
		\center
		\footnotesize
		\begin{threeparttable}
		  \caption{Top-1 Accuracy (\%) on CIFAR-100 Test Set}
		  \label{cifar-100_results}
		  \begin{tabular}{ccccccc}
			\toprule  
			Methods   & MobileNet \cite{mobilenet} & MobileNet V2 \cite{mobilenetv2} & ShuffleNet V2 \cite{shufflenetv2} & SqueezeNet \cite{squeezenet} & VGG-11 \cite{vgg} & VGG-13  \\
			\midrule
			ReLU      & $67.52 \pm 0.18$ & $68.61 \pm 0.32$ & $70.52 \pm 0.28$ & $70.37 \pm 0.39$ & $68.36 \pm 0.38$ & $72.57 \pm 0.35$ \\
			LReLU & $67.72 \pm 0.12$ & $69.29 \pm 0.17$ & $71.02 \pm 0.15$ & $70.13 \pm 0.31$ & $68.55 \pm 0.18$ & $72.35 \pm 0.22$ \\
			PReLU     & $67.84 \pm 0.46$ & $69.20 \pm 0.34$ & $70.28 \pm 0.29$ & $68.77 \pm 0.21$ & $66.57 \pm 0.50$ & $70.92 \pm 0.08$ \\
			Swish     & - & - & $72.24 \pm 0.08$ & $69.69 \pm 0.27$ & $68.03 \pm 0.08$ & $71.99 \pm 0.20$ \\
			SiLU      & $70.52 \pm 0.30$ & $69.97 \pm 0.11$ & $71.79 \pm 0.26$ & $70.62 \pm 0.17$ & $67.10 \pm 0.17$ & $71.17 \pm 0.38$ \\
			Mish      & $71.17 \pm 0.15$ & $70.20 \pm 0.22$ & $71.65 \pm 0.27$ & $70.55 \pm 0.21$ & $67.26 \pm 0.12$ & $71.59 \pm 0.24$ \\
			GELU      & $71.02 \pm 0.18$ & $71.24 \pm 0.20$ & $72.51 \pm 0.34$ & $70.56 \pm 0.33$ & $68.51 \pm 0.12$ & $72.41 \pm 0.25$ \\
			\midrule
			SGELU     & $\bm{71.80 \pm 0.36}$ & $\bm{71.81 \pm 0.19}$ & $\bm{73.05 \pm 0.09}$ & $70.38 \pm 0.47$ & $\bm{69.47 \pm 0.26}$ & $\bm{73.28 \pm 0.36}$ \\
			SSiLU     & $71.73 \pm 0.17$ & $71.33 \pm 0.30$ & $72.82 \pm 0.10$ & $\bm{71.13 \pm 0.27}$ & $68.86 \pm 0.13$ & $72.32 \pm 0.12$ \\
			SMish     & $71.62 \pm 0.17$ & $70.83 \pm 0.28$ & $72.37 \pm 0.28$ & $70.86 \pm 0.54$ & $68.76 \pm 0.09$ & $71.98 \pm 0.21$ \\
			\bottomrule 
		  \end{tabular}
		  \begin{tablenotes}
			\item [1] "$-$" indicates that the training with this activation function does not converge.
		  \end{tablenotes}
		\end{threeparttable}
	  \end{table}

	Table \ref{cifar-100_results} shows the experimental results. The experiments showed that all three activation functions we proposed have excellent performance. In deep learning networks other than SqueezeNet, SGELU outperforms GELU, increasing classification accuracy by about 0.2\% to 1.0\%. SSiLU and SMish both perform better on all networks used in the experiment than SiLU and Mish respectively. In all networks, the best method is our proposed method. SGELU performs the best, achieving the best classification accuracy in all deep learning networks except for SqueezeNet, and improving classification accuracy by about 1\% to 4.3\% compared to ReLU. SSiLU improves classification accuracy by 0.5\% to 4.2\% compared to ReLU, except for being 0.25\% lower than ReLU in VGG-13. SMish improves classification accuracy by 0.5\% to 4.2\% compared to ReLU, except for being 0.59\% lower than ReLU in VGG-13. Compared with other baseline activation functions, SGELU performs only 0.24\% lower than the best baseline method in SqueezeNet and improves performance by 0.2\% to 0.9\% over the best baseline method in other networks. SSiLU improves performance by 0.1\% to 0.6\% over the best baseline method in 5 networks except for being 0.25\% lower than the best baseline method ReLU in VGG-13. SMish improves performance by 0.2\% to 0.5\% over the best baseline method in 3 networks, while it is 0.41\% lower on MobileNet V2, 0.14\% lower on ShuffleNet V2, and 0.59\% lower than the best baseline method ReLU on VGG-13.

	In summary, our method combines a non-monotonic activation function with ReLU, and the performance of the combined activation function is significantly improved compared to the original non-monotonic activation function and ReLU. The extent of improvement is related to the performance of the original non-monotonic activation function: in this experiment, the classification accuracy rankings of the three non-monotonic activation functions are $\text{GELU} > \text{SiLU} > \text{Mish}$, and thus the rankings for our proposed activation functions are $\text{SGELU} > \text{SSiLU} > \text{SMish}$. SGELU outperforms other baseline activation functions in all networks.

	\section{Conclusion}
	In this paper, we investigated the properties of ReLU and some non-monotonic activation functions. ReLU can activate positive features without distortion, while non-monotonic activation functions provide nonlinearity with non-monotonic negative parts. We propose a method to combine the advantages of both, retaining the negative nonlinearity and replacing the positive part with a linear function. This allows the propagation of negative signals while activating positive features without distortion. Based on this method, we propose three activation functions, SGELU, SSiLU and SMish, which are combinations of the negative part of GELU, SiLU, and Mish respectively and the positive part of ReLU. Our proposed activation functions demonstrate excellent performance in CIFAR-100 image classification experiments, with SGELU achieving the best test accuracy. 

	\bibliographystyle{elsarticle-num}
	\bibliography{cite}

\end{document}